\documentclass{article}

\usepackage{microtype}
\usepackage{graphicx}
\usepackage{subfigure}
\usepackage{booktabs} 
\usepackage[dvipsnames]{xcolor}  
\usepackage{inputenc}
\usepackage{enumitem}
\usepackage{amsmath,siunitx}
\usepackage{svg}
\usepackage{psfrag,auto-pst-pdf}

\usepackage{hyperref}
\usepackage[hyphenbreaks]{breakurl}

\newcommand\etal{\textit{et. al.}}

\usepackage[framemethod=tikz,hidealllines=true,backgroundcolor=yellow!19]{mdframed}
\usetikzlibrary{arrows}

\usepackage[backend=biber,style=numeric-comp,giveninits=true,sorting=none]{biblatex}
\DeclareNameAlias{default}{last-first}
\DeclareBibliographyCategory{important}
\addtocategory{important}{%
tarapore2020sparse,%
das_tarmac:_2019,%
ShumeyeLakew.2020%
}
\renewbibmacro*{begentry}{\ifcategory{important}{\textbullet\addspace}{}}

\renewbibmacro*{doi+eprint+url}{%
    \printfield{doi}%
    \newunit\newblock%
    \iftoggle{bbx:eprint}{%
        \usebibmacro{eprint}%
    }{}%
    \newunit\newblock%
    \iffieldundef{doi}{%
        \usebibmacro{url+urldate}}%
        {}%
}

\usepackage[accepted]{icml2021}

\icmltitlerunning{A Review of Communications in Multi-Robot Systems}

\begin{document}

\twocolumn[
\icmltitle{A Critical Review of Communications in Multi-Robot Systems} 




\icmlsetsymbol{equal}{*}

\begin{icmlauthorlist}
\icmlauthor{Jennifer Gielis}{equal,cam}
\icmlauthor{Ajay Shankar}{equal,cam}
\icmlauthor{Amanda Prorok}{cam}
\end{icmlauthorlist}

\icmlaffiliation{cam}{Department of Computer Science and Technology, University of Cambridge, United Kingdom}

\icmlcorrespondingauthor{Jennifer Gielis}{jag233@cl.cam.ac.uk}
\icmlcorrespondingauthor{Ajay Shankar}{as3233@cam.ac.uk}

\icmlkeywords{Machine Learning, ICML}

\vskip 0.3in]



\printAffiliationsAndNotice{\icmlEqualContribution} 

\begin{abstract}

\textbf{Purpose of Review.}
This review summarizes the broad roles that communication formats and technologies have played in enabling multi-robot systems.
We approach this field from two perspectives: of robotic applications that \textit{need} communication capabilities in order to accomplish tasks, and of networking technologies that have enabled newer and more advanced multi-robot systems.
\newline{}\textbf{Recent Findings.}
Through this review, we identify a dearth of work that holistically tackles the problem of co-design and co-optimization of robots and the networks they employ.
We also highlight the role that data-driven and machine-learning approaches play in evolving communication 
pipelines for multi-robot systems.
In particular, we refer to recent work that diverges from hand-designed communication patterns, and also discuss the `sim-to-real' gap in this context.
\newline{}\textbf{Summary.}
We present a critical view of the way robotic algorithms and their networking systems have evolved, and make the case for a more synergistic approach.
Finally, we also identify four broad \textit{Open Problems} for research and development, while offering a data-driven perspective for solving some of them.
\end{abstract}





\section{Introduction}
\label{sec:introduction}


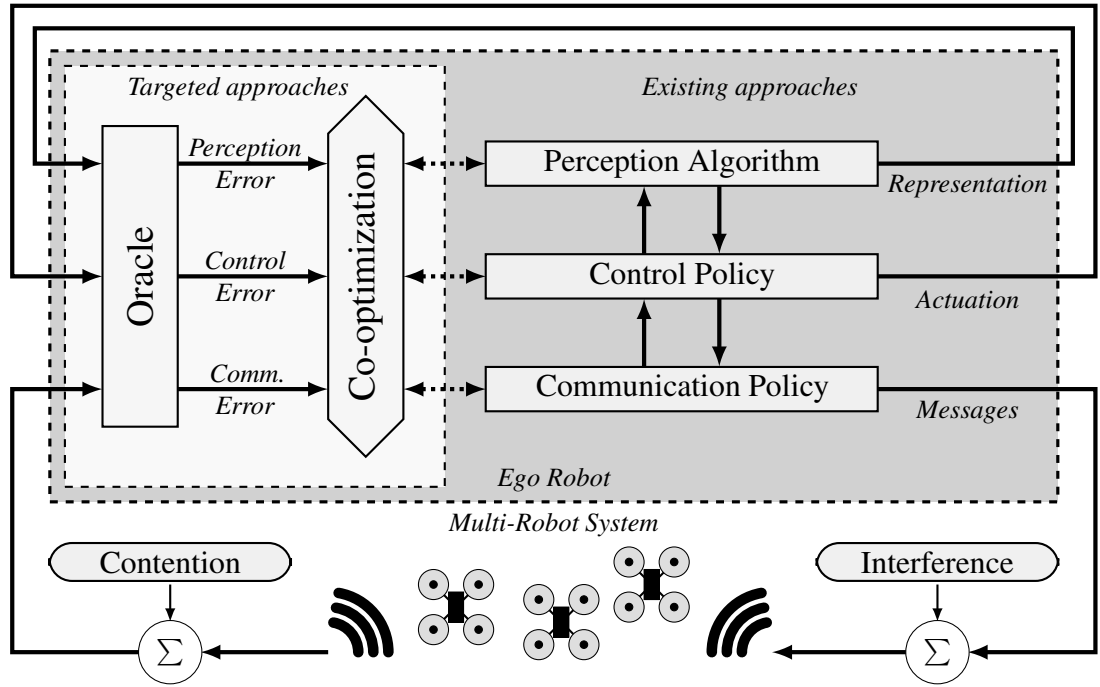
\begin{figure*}
    \centering
\begin{tikzpicture}[font=\normalsize]
	\definecolor{refcolor}{HTML}{bdbdbd}
	\definecolor{boxcolor}{HTML}{f0f0f0}
	\tikzstyle{tbox} = [draw,thick,text width=4cm,align=flush center,fill=boxcolor,inner sep=3pt];
	\tikzstyle{cbox} = [text width=3.8cm,align=flush center,fill=refcolor!60,inner sep=3pt];
	 \tikzstyle{year_label} = [];
	\tikzstyle{year_tick} = [thick];
	\tikzstyle{base_box_con} = [thick,smooth,tension=.9,latex-];
	
	\draw[very thick, dashed, fill=refcolor!70]  (-9,1) rectangle (4.4,-5);
	\draw[thick, dashed, fill=refcolor!10]  (-8.8,0.8) rectangle (-3.75,-4.8);
	
	\draw[ultra thick, -latex] (-5.3,-7) -- (-7,-7);
	
	\draw[ultra thick, -latex] (0.9,-0.5) -- (4.6,-0.5) -- (4.6,1.3) -- (-9.2,1.3) -- (-9.2,-0.5) -- (-8.3,-0.5);
	\draw[ultra thick, -latex] (0.9,-2) -- (4.9,-2) -- (4.9,1.6) -- (-9.5,1.6) -- (-9.5,-2) -- (-8.3,-2);
	\draw[ultra thick, -latex] (0.9,-3.5) -- (4.9,-3.5) -- (4.9,-7) -- (3.2,-7);
	\draw[ultra thick, -latex] (-7.8,-7) -- (-9.5,-7) -- (-9.5,-3.5) -- (-8.3,-3.5);

	\draw[ultra thick, -latex] (2.4,-7) -- (0.6,-7);

	\draw[thick, -latex] (2.8,-6.1) -- (2.8,-6.6);
	\draw[thick, -latex] (-7.4,-6.1) -- (-7.4,-6.6);

	\draw[ultra thick, -latex] (-0.1,-0.8) -- (-0.1,-1.7);
	\draw[ultra thick, -latex] (-0.1,-2.3) -- (-0.1,-3.2);
	\draw[ultra thick, -latex] (-1.1,-3.2) -- (-1.1,-2.3);
	\draw[ultra thick, -latex] (-1.1,-1.7) -- (-1.1,-0.8);
	
	\draw[ultra thick, -latex] (-7.3,-0.5) -- (-5.3,-0.5);
	\draw[ultra thick, -latex] (-7.3,-2) -- (-5.3,-2);
	\draw[ultra thick, -latex] (-7.3,-3.5) -- (-5.3,-3.5);
	
	\draw[ultra thick, dotted, latex-latex] (-4.3,-0.5) -- (-3.2,-0.5);
	\draw[ultra thick, dotted, latex-latex] (-4.3,-2) -- (-3.2,-2);
	\draw[ultra thick, dotted, latex-latex] (-4.3,-3.5) -- (-3.2,-3.5);

	\node[text width=3cm,,align=flush center] at (-6.4,-0.5) {\emph{Perception\\ Error}};
	\node[text width=3cm,,align=flush center] at (-6.4,-2) {\emph{Control\\ Error}};
	\node[text width=3cm,,align=flush center] at (-6.4,-3.5) {\emph{Comm.\\ Error}};
	
	\node at (3.2,-0.8) {\emph{Representation}};
	\node at (3.2,-2.3) {\emph{Actuation}};
	\node at (3.2,-3.8) {\emph{Messages}};
	
	\node[tbox,text width=5cm] at (-0.6,-2) {\large Control Policy};
	\node[tbox,text width=5cm] at (-0.6,-3.5) {\large Communication Policy};
	\node [tbox,text width=5cm] at (-0.6,-0.5) {\large Perception Algorithm};
	\node [tbox,rounded corners=.3cm,text width=3cm] at (2.8,-5.8) {\large Interference};
	
	\node [tbox,rounded corners=.3cm,text width=3cm] at (-7.4,-5.8) {\large Contention};
	
	\draw[thick,fill=boxcolor!60] (-4.8,0.4) node (v6) {} -- (-4.3,-0.1) -- (-4.3,-3.9) -- (-4.8,-4.4) -- (-5.3,-3.9) -- (-5.3,-0.1) -- cycle;
	\node[rotate=90] at (-4.8,-2) {\Large Co-optimization};
	
	\draw[thick,fill=boxcolor!60]  (-8.3,0) rectangle (-7.3,-4);
	\node[rotate=90] at (-7.8,-2) {\Large Oracle};

	\node[draw, circle, fill=white] at (-7.4,-7) {$\sum$};
	\node[draw, circle, fill=white] at (2.8,-7) {$\sum$};
	
	\node at (-2.3,-4.7) {\emph{Ego Robot}};
	\node at (0.3,0.5) {\emph{Existing approaches}};
	\node at (-6.5,0.5) {\emph{Targeted approaches}};
	\node at (-2.3,-5.3) {\emph{Multi-Robot System}};

	\tikzstyle{wifibar} = [line width=4pt, line cap=round];
	\tikzstyle{droneprop} = [fill=refcolor!50, fill opacity=100];
	\draw[wifibar] (0.2,-7) arc (180:100:0.4);
	\draw[wifibar] (0,-7) arc (180:100:0.6);
	\draw[wifibar](-0.2,-7) arc (180:100:0.8);
	\draw[wifibar] (-4.9,-7) arc (0:80:0.4);
	\draw[wifibar] (-4.7,-7) arc (0:80:0.6);
	\draw[wifibar] (-4.5,-7) arc (0:80:0.8);

	\draw[thick] (-2.5,-6.3) node (v4) {} -- (-1.9,-6.9) node (v2) {};
	\draw[thick] (-2.5,-6.9) node (v1) {} -- (-1.9,-6.3) node (v3) {};
	\draw[fill]  (-2.3,-6.4) rectangle (-2.1,-6.8);
	\draw[droneprop]  (v1) ellipse (0.2 and 0.2);
	\draw[droneprop]  (v2) node (v5) {} ellipse (0.2 and 0.2);
	\draw[droneprop]  (v3) ellipse (0.2 and 0.2);
	\draw[droneprop]  (v4) ellipse (0.2 and 0.2);
	\draw[fill] (v1) ellipse (0.05 and 0.05);
	\draw[fill] (v2) ellipse (0.05 and 0.05);
	\draw[fill] (v3) ellipse (0.05 and 0.05);
	\draw[fill] (v4) ellipse (0.05 and 0.05);
	
	\draw[thick] (-1.3,-5.8) node (v4) {} -- (-0.7,-6.4) node (v2) {};
	\draw[thick] (-1.3,-6.4) node (v1) {} -- (-0.7,-5.8) node (v3) {};
	\draw[fill]  (-1.1,-5.9) rectangle (-0.9,-6.3);
	\draw[droneprop]  (v1) ellipse (0.2 and 0.2);
	\draw[droneprop]  (v2) node (v5) {} ellipse (0.2 and 0.2);
	\draw[droneprop]  (v3) ellipse (0.2 and 0.2);
	\draw[droneprop]  (v4) ellipse (0.2 and 0.2);
	\draw[fill] (v1) ellipse (0.05 and 0.05);
	\draw[fill] (v2) ellipse (0.05 and 0.05);
	\draw[fill] (v3) ellipse (0.05 and 0.05);
	\draw[fill] (v4) ellipse (0.05 and 0.05);
	
	\draw[thick] (-3.9,-6.1) node (v4) {} -- (-3.3,-6.7) node (v2) {};
	\draw[thick] (-3.9,-6.7) node (v1) {} -- (-3.3,-6.1) node (v3) {};
	\draw[fill]  (-3.7,-6.2) rectangle (-3.5,-6.6);
	\draw[droneprop]  (v1) ellipse (0.2 and 0.2);
	\draw[droneprop]  (v2) node (v5) {} ellipse (0.2 and 0.2);
	\draw[droneprop]  (v3) ellipse (0.2 and 0.2);
	\draw[droneprop]  (v4) ellipse (0.2 and 0.2);
	\draw[fill] (v1) ellipse (0.05 and 0.05);
	\draw[fill] (v2) ellipse (0.05 and 0.05);
	\draw[fill] (v3) ellipse (0.05 and 0.05);
	\draw[fill] (v4) ellipse (0.05 and 0.05);
\end{tikzpicture}
    \caption{A flow diagram showing how perception, control and communications typically interact in a multi-robot system. On the left is a proposed future approach to co-optimization, discussed in Section \ref{sec:future}, which now admits communications and its confounding factors as integral to any co-optimization strategy designed for a multi-robot system.}
    \label{fig:flowdiagram}
\end{figure*}

{The use of} {multiple, \textit{connected} robots {in the place of} individually {uncommunicative} robots} provides evident gains by facilitating the inter-robot coordination that allows for work distribution, spatial coverage, and specialization.
An increasing variety of applications leverage such networks of robots, including  
logistics~\cite{tilley_Automation_2017,kamagaew_Concept_2011},
resource distribution~\cite{enright_Optimization_2011b, ma_Lifelong_2017a},
transport systems~\cite{hyldmar_Fleet_2019b, dressler_Intervehicle_2014a, ferreira_Selforganized_2010a},
manufacturing~\cite{cherubini_Collaborative_2016}, and agriculture~\cite{noguchi_Robot_2011, albani_Monitoring_2017}.

These applications depend on an orchestration of robots over time and space that allows them to jointly work towards common higher-order goals, to deconflict individual actions in shared environments, and to share information in distributed computing schemes.
Communication and the mutual exchange of information (state and control) are key to facilitating such interactions.

Early work in the multi-robot domain drew from nature-inspired paradigms~\cite{bonabeau1999swarm}, 
and consequently focused on devising collective behaviors that depended purely on \textit{local} interactions of robots in close proximity~\cite{nagpal2003organizing}. A variety of transmission media (e.g., infrared) are used for such near-field communication schemes~\cite{rubenstein2014programmable, pugh2006relative}. 
Other nature-inspired work built on \textit{implicit} communication and self-organization through stigmergy, by which robots coordinate indirectly through traces left in the environment~\cite{beckers2000fom}. 
The benefits of such peer-to-peer \textit{decentralized} communication paradigms are manyfold, in particular due to their inherent robustness and scalability~\cite{brambilla2013swarm}.
\textit{Centralized} radio-based communication architectures have become increasingly popular in various instances, especially when the task requires performance guarantees; representative applications include product pickup and delivery~\cite{grippa_Drone_2019a}, item retrieval in warehouses~\cite{enright_Optimization_2011b}, and mobility-on-demand services~\cite{spieser2016shared}. 
Improvements in communication technologies, both hardware and software, have furthered more data-intensive applications, such as cloud robotics~\cite{chinchali2021network, kehoe2015survey}.

Explicit communication methods generally assume that robots can broadcast information within a local neighborhood that comprises of tens to hundreds of individuals, or that a fixed network infrastructure is available.
Yet, in reality, densely populated workspaces adversely affect communication capabilities because of practical contention over channel bandwidth and airtime~\cite{gielis2021improving}.
Such networks are additionally burdened by clutter that can induce signal fading, leading to a drastic decrease in the expected communication range. 
This problem is compounded by the need for real-time transmission requirements in highly dynamic robot networks. Indeed, topologies and capabilities \textit{demanded} by robotic applications are practically hostile to radio performance (because these radio networks were not initially designed with robotics in mind).
As a consequence, the vast majority of robot applications are designed to merely work around available network technologies, and optimize their performance within the given constraints.



Our review is motivated by a lack of studies that provide a high-level overview of the interplay between communication networks and their role in robotic applications.
Figure~\ref{fig:flowdiagram}, graphically demonstrates the typical architecture of a multi-robot control scheme.
As mentioned prior, robot control algorithms generally do not actively employ the output of the communications network being simulated.
The result is a wide array of optimizations that work in favor of the network, but often not for autonomy, or vice-versa.
Hence, we argue that a better co-optimization scheme (illustrated on the left in Figure~\ref{fig:flowdiagram}) will consider all aspects of the architecture simultaneously.

In this survey, we capture a variety of network architectures and technologies, and a variety of multi-robot applications that employ them.
A careful choice of communications architecture, medium and algorithm are key to ensuring that a given robot task can be completed.
Therefore, we will also explore some of the newer approaches that consider bypassing such hand-crafted selections, and attempt to model inter-robot communications in a data-driven fashion.

\section{Factors Influencing Robot Network Design}
\label{sec:factors}

Choosing an adequate communications architecture, medium, and algorithm are key to ensuring desired robot performance. In the following, we distill the factors that influence the robot network design choices. We elaborate upon them in the following four categories, \textit{(i)} application, \textit{(ii)} robot, \textit{(iii)} algorithm, and \textit{(iv)} environment, and give an illustrative example for each.

\textbf{The application:}
The application defines \textit{what the shared information is for}, and \textit{how the robots need to interact} to solve the problem at hand. \emph{Examples}: Real-world applications such as in environmental monitoring and agriculture require groups of robots to act over large distances (often operating with robots separated by $\sim$1000x body lengths). Such sparsely distributed robot systems, hence, necessitate networking capabilities that can span larger spaces~\cite{tarapore2020sparse}. 
Other applications, such as cooperative driving~\cite{hyldmar_Fleet_2019b}, formation control~\cite{preiss2017crazyswarm}, and flocking~\cite{Tolstaya19-Flocking} require uninterrupted, situated, close-range communication for tight inter-robot coordination and control.



\textbf{The robot:} The robot (and the physical hardware) define local \textit{constraints on the frequency and format} of information to be transmitted and received. \textit{Examples:} A quadrotor that uses state information for local stability control requires an update frequency in the order of several hundred Hertz; while on-board IMUs can provide the necessary information for body stabilization, extrinsic pose estimates are still required for tasks, and must be received at relatively high rates (e.g., \SI{100}{Hz})~\cite{preiss2017crazyswarm, kushleyev2013towards}. 
Lack of reliable updates naturally poses a significant risk to tasks that require tight coordination, such as outdoor flocking and formation control; while sparse outdoor flight has been demonstrated in a team of 30 drones~\cite{vasarhelyi2018optimized}, there is a dearth of results on \textit{dense and agile} outdoor flight.
Moreover, in GPS-denied environments, robots resort to on-board sensing,
and consequently, require dependable inter-robot communication to achieve group behavior.



\textbf{The algorithm:} The algorithm connects the \textit{application} to the \textit{robot}, and essentially sets conditions on the nature of information that needs to be received (e.g., global or local), and when (e.g., asynchronously or synchronously, and how often). \textit{Examples:} In allocation problems, the optimization objective is often global, and to achieve optimality, we deploy centralized algorithms that collect all robot-to-task assignment costs (e.g., expected travel times) to determine the optimal assignment (e.g., by running the Hungarian algorithm)~\cite{prorok_Redundant_2019a, khamis2015multi}. Similarly, multi-robot path planning has an optimal solution (for both makespan and flowtime objectives), but only when the computational unit has access to full system information~\cite{yu_Optimal_2015}. In the absence of full observability, robots need to resort to locally available knowledge. In \textit{decoupled} prioritized path planning, robots communicate to mutually deconflict their path plans in time-space~\cite{wu_MultiRobot_2019, cap_Asynchronous_2013, desaraju_Decentralized_2012}. Each time a robot's plan changes, its robot neighborhood changes, or a new conflict arises, the deconfliction process restarts.




\textbf{The environment:}
The environment defines \textit{under what conditions} shared information is delivered. \textit{Examples:} Are the robots operating indoors or outdoors, or both~\cite{dong2015distributed}?
Does the workspace afford a fixed (and possibly centralized) communications infrastructure, or must we instead rely on ad-hoc networking? 
Is the environment cluttered with obstacles that interfere with wireless signals? What medium can we use, e.g., are the robots operating in air, under water, or in space?
What legal jurisdictions regulate the communication infrastructure?
And finally, is the communication channel safe, or can it be spoofed~\cite{gil_Guaranteeing_2017}, or robots attacked~\cite{saulnier_Resilient_2017a,guerrero-bonilla_Dense_2020}?



\begin{figure*}
    \centering
\begin{tikzpicture}[font=\normalsize]
\definecolor{refcolor}{HTML}{bdbdbd}
\definecolor{boxcolor}{HTML}{f0f0f0}
    \fill [left color=gray!80,right color=gray!0,draw=white] 
    (1,0.15) .. controls (9.6,0.4) and (10.1,0.5) .. (15,4.4)
    --(15,-4.4)
    (15 ,-4.4) .. controls (10.1,-0.5) and (9.6,-0.5) .. (1,-0.15)
    --(1, 0.15);
    \draw [ultra thick,-triangle 60](1,0) -- (16.5,0);
    
    \tikzstyle{tbox} = [draw,thick,text width=3.5cm,align=flush center,fill=boxcolor!60,inner sep=3pt];
    \tikzstyle{cbox} = [text width=3.4cm,align=flush center,fill=refcolor!60,inner sep=3pt];
    \tikzstyle{year_label} = [];
    \tikzstyle{year_tick} = [thick];
    \tikzstyle{base_box_con} = [thick,smooth,tension=.9,latex-];
    
    \draw [base_box_con] plot coordinates {(1.6,0) (1.5,-1) (2.4,-1.8)};
    \node [cbox] at (2.3,-3.6) {\cite{stojanovic1996recent}};
    \node [tbox] at (2.3,-2.6) {Acoustic-LAN w/ QPSK/QAM modems\\[5pt]
    Underwater robots};
    
    \draw [base_box_con] plot coordinates {(1.7,0) (1.4,2) (1.7,4)};
    \node [cbox, text width=4.4cm] at (2.7, 3.05) {\cite{Wang.1994,Cao.1997}};
    \node [tbox, text width=4.5cm] at (2.7,4.05) {\emph{Bespoke radio data formats}\\[5pt]Nascent multi robot control\\ Human in the loop};
    
    \node [year_label] at (2.5,-0.5) {1997};
    \node (1997base) at (2.5,0) {};
    \draw [year_tick](2.5,-0.2) -- (2.5,0.2);
    \draw [base_box_con] plot coordinates {(1997base) (2.1,1) (2.4,1.8)};
    \node [cbox, text width=2.9cm] at (3.8,1) {\cite{IEEE.1997}};
    \node (qwe) [tbox, text width=3cm] at (3.8,2) {\emph{802.11}\\[5pt] Local Wireless data network standards };
    
    \node [year_label] at (3.5,0.5) {1999};
    \node (1999base) at (3.5,0) {};
    \draw [year_tick](3.5,-0.2) -- (3.5,0.2);
    \draw [base_box_con] plot coordinates {(1999base) (4.9,-1.2) (6.4,-1.8)};
    \node [cbox] at (6.25,-3.8) {\cite{Luu.2007}};
    \node [tbox] at (6.25,-2.15) {\emph{802.11a}\\ 5GHz, Orthogonal frequency division (OFDM)\\[5pt] Consumer Hardware\\Sensor-speed (54Mbps)};
    
    \node (2004base) at (6,0) {};
    \node [year_label] at (6,-0.5) {2004};
    \draw [year_tick](6,-0.2) -- (6,0.2);
    \draw [base_box_con] plot coordinates {(2004base) (6.7,1.1) (6.5,2.6)};
    \node [cbox] at (7.7,2.75) {\cite{Houda.2015}};
    \node [tbox] at (7.7,3.75) {\emph{802.15.4}, \emph{ZigBee}\\[5pt] Mesh networks\\ Low power radio};

    \node (2010base) at (9,0) {};
    \node [year_label] at (9,0.5) {2010};
    \draw [year_tick](9,-0.2) -- (9,0.2);
    \draw [base_box_con] plot coordinates {(8.75,0) (9,-1.3) (9.9,-2.6)};
    \node [cbox] at (10.2,-3.8) {\cite{Naik.2019}};
    \node [tbox] at (10.2,-2.35) {\emph{802.11n/p}\\MIMO\\[5pt]Automotive focus\\ Doppler shift handling (up to 250km/h)};
    
    \node (2015base) at (11.5,0) {};
    \node [year_label] at (11.5,-0.5) {2015};
    \draw [year_tick](11.5,-0.2) -- (11.5,0.2);
    \draw [base_box_con] plot[smooth, tension=.7] coordinates {(2015base) (10.5,0.5) (10,1.6)};
    \node [cbox] at (9.5,1) {\cite{Erturk.2019}};
    \node [tbox, text width=3.5cm] at (9.5,1.8) {\emph{LoRaWAN}\\[5pt]Large scale low power};
    
    \node (2019base) at (13.5,0) {};
    \node [year_label] at (13.5,-0.5) {2019};
    \draw [year_tick](13.5,-0.2) -- (13.5,0.2);
    \draw [base_box_con] plot coordinates {(2019base) (12.3,1.2) (11.9,2.6)};
    \node [cbox] at (11.8,2.7) {\cite{Feng.2021}};
    \node [tbox] at (11.8,3.9) {\emph{5G URLLC, 802.11ax}\\Ultra-high throughput\\[3pt] Cloud/edge robotics\\Low latency cellular};

    \node (2022base) at (15,0) {};
    \node [year_label] at (15,-0.5) {2022};
    \draw [year_tick](15,-0.2) -- (15,0.2);
    

    \draw [base_box_con] plot coordinates {(16.2,0) (15.5,-0.5) (16,-1.5)};
    \node [tbox,text width=5cm] at (15,-2.5) {
    \emph{Future Network Capabilities}\\
    Large scale low-latency networks\\
    Contention resilience for ad-hoc\\[5pt]
    \emph{Future Robotic Applications}\\
    Large scale swarm deployment\\
    Reliable real world behavior};
    
    \draw [base_box_con] plot coordinates {(13.8,0) (13.5,1.25) (14.4,2.5)};
    \node [cbox,text width=3.5cm] at (15.75,1.9) {\cite{electronics9020351}};
    \node [tbox,text width=3.5cm] at (15.75,3.3) {
    \emph{802.15.3 UWB, 802.11az}\\[5pt]
    Increasingly accurate localization (802.11az/5G \textless1m, 802.15.3 \textless10cm)};

\end{tikzpicture}
    \caption{An abridged timeline showing some key wireless communication mechanisms for robots. The shaded area represents the magnitude of theoretical capabilities (bandwidth or inverse latency), which have been increasing super-linearly since 2010. Boxes show communications standards or technologies, and their specific properties that are useful in the multi-robot control space.}
    \label{fig:timeline}
\end{figure*}
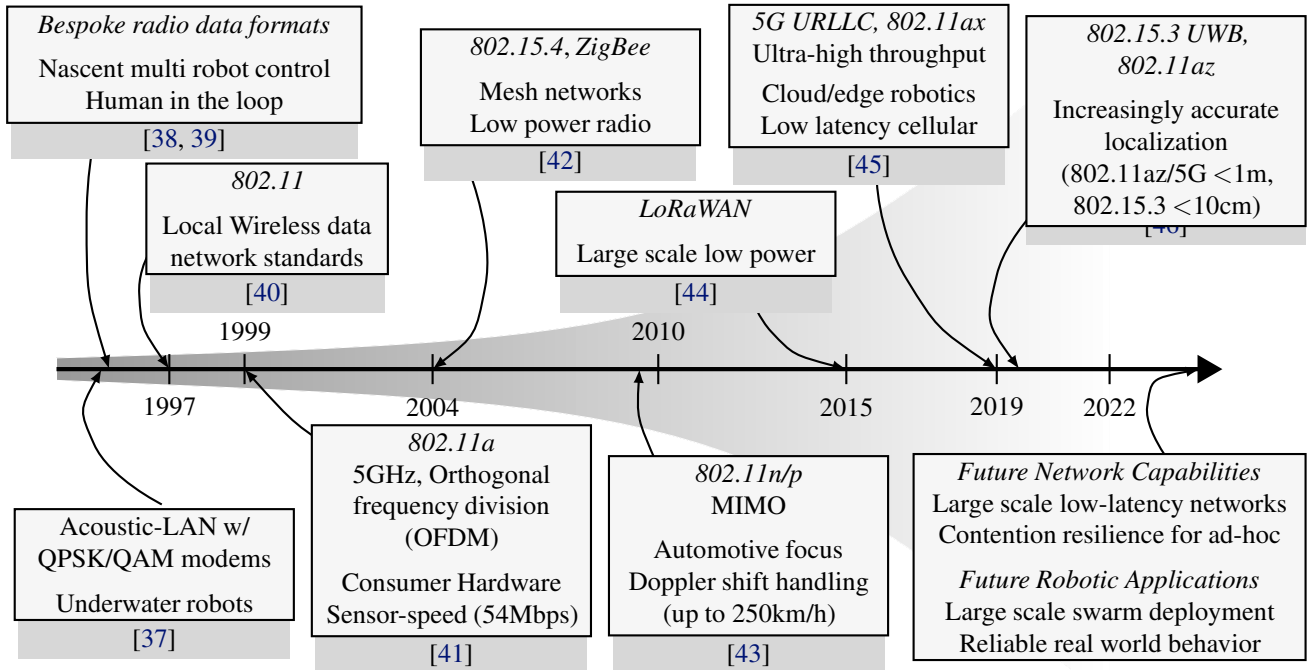


\section{Communication Schemes}
\label{sec:comms}
In this section we discuss multi-robot communication from the perspective of the underlying communications technologies, focusing upon the challenges, limitations and optimizations that are relevant in multi-robot system networks. Fig.~\ref{fig:timeline} shows a timeline with key wireless communication mechanisms, and some representative multi-robot applications that they enabled for the first time.

\subsection{Challenges}
\textbf{Synchronicity.} Specifying robotic data flows is often the first consideration in discussing the challenges of a wireless data protocol. 
For example, it is often implicitly assumed that multi-robot control algorithms are executed synchronously by every participant~\cite{vandenberg_Reciprocal_2008}. This introduces a hard timing constraint on the maximum allowable delay in message delivery between those participants.
This is, however, a feature that commonly deployed communications protocols are not designed to meet, with `best-effort' message delivery being the standard paradigm~\cite{Koutsiamanis.2018}.

\textbf{Dynamic topologies}. Hard timing constraints are often exacerbated by highly connected communications topologies that are dynamic, where a robot must communicate its status with many different (or sometimes every) participating robot(s). This can lead to a high degree of contention for radio resources since many messages may need to be sent at every control loop. While there are schemes that aim to minimize redundant data transmissions (see Section~\ref{sec:comms_aware}), it remains true that, as multi-robot networks increase in scale, communications technologies must be selected and designed specifically to manage the dynamics of the application~\cite{Fan.2018}, something that is generally overlooked in robotic networks today.

\textbf{Message frequency.} Bandwidth is often employed as a metric to specify the demands on a communications link~\cite{Sarr.2008}. However, this is often an insufficient characterization by itself, since the underlying technology may have significant overheads per message, and robot teams often depend on low-latency messaging as well. This is particularly true for ad-hoc networks where there is no central entity enforcing message scheduling, to the extent that many communications protocols will not approach their rated bandwidths in highly connected ad-hoc topologies, where overheads such as contention dominate radio resource consumption~\cite{gielis2021improving}.

\textbf{Connectivity.}
Since a greater connectivity range implies reachability and information exchange with more robots, it has an obvious impact on the overall messaging rate any specific robot must handle.
The geographical density of robots must be considered in the discussion of range, and there are two key factors of interest. Firstly, as ranges increase, radio-based links are more prone to fading and interference~\cite{Yan.2019} even when transmission power is commensurately increased. Secondly, robot control algorithms often assume a fixed range~\cite{Khodayimehr.2019}, which may result in greater inter-connectivity, and sometimes increased messaging rates in dense scenarios.

\textbf{Dynamic routing.} The case where the required range of communication exceeds the underlying capabilities of the radio hardware onboard must also be considered, as this implies a mesh-type network where a message must traverse multiple robots (network nodes).This first requires planning the robots' paths (discussed in Section~\ref{subsec:comm-aware-planning}), before accounting for the computational and protocol overheads of robots processing messages other than their own. Then, the problem is that of message routing decisions and dynamic topologies. The routing decision problem is generally central to ad-hoc mesh networks; this is only made more challenging by the potential for rapid shifts in communications topology, especially in highly mobile, or large scale robotic scenarios~\cite{Gupta.2016}. Hard timing guarantees, in the range required by robotic control, are not currently available at non-trivial scales (especially with multi-hop routing over dynamic topologies), though some attempts have been made in this direction~\cite{Deng.2021}.

\textbf{Operational environment.} Robotic networks will invariably be required to operate in environments with external noise and interference, which cause unpredictable impacts on link quality. This informs the selection of communications protocols, since some protocols operate in a licensed spectrum with reduced external interference, or are otherwise less prone to external noise due to atmospheric attenuation (60GHz). Doppler shift requires similar consideration, because many communications technologies fail at high relative velocities. Generally speaking, protocols that depend upon fine-grained frequency division multiplexing are more prone to Doppler related errors~\cite{Elmezughi.2021c}, and such schemes are often used in high bandwidth techniques.

\subsection{Communications Scheme Selection}
\label{subsec:comms_protocols}
Despite considerable research interest, there are no current wireless data standards explicitly designed for exchanging information between autonomous robots~\cite{IEEE.2022}. Currently deployed robot-to-robot networks (such as ~\cite{vasarhelyi2018optimized}) depend upon more generic wireless data networking standards which are not typically optimized for the challenges discussed above. In the absence of a specific standard, we will discuss the strengths and weaknesses of existing technologies for the multi-robot control application.

Ad-hoc networks map well to the communications patterns required of decentralized robotic control, and the most relevant for this survey are Mobile Ad-Hoc Networks (MANETs) which deal with the problem of facilitating communication between mobile nodes without coordination from infrastructure ~\cite{D.Ramphull.2021}. More specific forms of interest are Vehicular Ad-Hoc Networks (VANETs) ~\cite{AlHeety.2020} and Flying Ad-hoc networks (FANETS)~\cite{Srivastava.2021}, where the former generally deal with automotive use cases and the latter drones and UAVs, and these are more exposed to dynamic conditions that are expected from robotic ad-hoc networks. Local area networking technologies are well suited for ad-hoc networking.

In contrast with ad-hoc operation, infrastructure networks map more closely to a centralized robotic control, where communications patterns are more similar to traditional bandwidth-focused networking applications. Despite this, hard latency requirements and rapid robot movement require specific mechanisms at the protocol level, which are not common for either Cellular or local area standards.

\subsubsection{Local Area Networks}
The IEEE 802.11 protocol suite, commonly known as `wi-fi', is frequently used due to abundant hardware availability, IP networking interoperability, high data rates and license-free operation. It is also capable of both infrastructure and ad-hoc operations, which simplifies deployment from laboratory environments into the real world. The failures of 802.11 become apparent during such deployment processes~\cite{Tahir.2021, HameedMir.2014}, because larger ranges, robot counts and velocity-induced Doppler shift cause lower message delivery rates than are seen in static 802.11 deployments. 802.11p, and its successor 802.11bd, have both introduced specific modifications to the \textit{physical} layer ~\cite{Naik.2019} to more robustly handle both range and Doppler-induced problems for VANET use-cases, which potentially transfer to robotic control as well.

IEEE 802.15.4 has been commonly used as the basis for a number of different higher-layer protocols, including ZigBee.
It has been deployed in the context of wireless sensor networks and multi-robot systems~\cite{Houda.2015} due to hardware availability, license-free operation, low power usage, and flexible communications models that permit both IP-based and more simplified serial-like messaging. The major drawbacks are relatively low range and data rates. LoRaWAN~\cite{Erturk.2019} is an attractive alternative that maintains the positive aspects of 802.15.4 for the robotic use cases, but with a focus upon long range transmission (up to 16km) and a physical layer that is resilient to Doppler errors; however, its communications model is infrastructure based. 

Both 802.11's and 802.15's underlying dependence upon the CSMA/CA collision avoidance scheme allows for the minimization of contention related losses without an authoritative central scheduler~\cite{Ziouva.2002}; however, they have unbounded maximum latency on message delivery, and reduced message delivery rates with higher numbers of robots on the network. LoRaWAN uses a pure ALOHA protocol mechanism~\cite{Adelantado.2017}, and therefore scales even more poorly than the IEEE schemes. These characteristics make these protocols unsuitable for deployment on robots without modifications; fortunately, there are techniques proposed in the literature to help make these technologies more scalable~\cite{Shahin.2018, Petrosky.2019}.

All of the protocols mentioned within this section share similar routing problems when it comes to highly dynamic topologies, in that they depend upon a network-layer routing mechanism to direct traffic without reliably converged information about the current disposition of other nodes. This issue is well covered by \cite{ShumeyeLakew.2020}, which categorizes and surveys many of the different approaches towards this routing problem. \cite{Hentati.2020} includes the routing issue amongst a general survey of the issues in UAV networking.

\subsubsection{Cellular Networks}
Cellular networks avoid many of the problems encountered by local networking standards by making use of a nearly universal infrastructure-based communications model, licensed radio spectrum access, as well as economics of scale, all of which make extremely complex base station hardware and protocols commonplace. The centralized message scheduler and sophisticated radio resource management~\cite{Chataut.2020, Liu.2016} are significantly more scalable than typical ad-hoc networks. These characteristics appear to be a good fit for robot control, however, the financial cost of network access, coupled with limited flexibility in logical network configuration have limited robotic deployments outside controlled environments.

For decentralized robot control, peer to peer traffic is routed through the infrastructure, inducing a minimum latency overhead~\cite{Chen.2018b} that could exceed timing constraints. 4G in particular has an access latency on the order of 50ms~\cite{Tahir.2021}. Additionally, cellular standards are naturally dependent upon the presence of infrastructure, which cannot always be assumed. Furthermore, the radiation pattern of cellular networks is typically setup assuming ground based users, and so aerial robots could experience degraded performance due to leaving the vertical coverage of cell antennas~\cite{Zeng.2019}.

4G LTE supports direct Device-to-Device (D2D) modes that permit devices to communicate with each other in a local region by reserving some subset of radio resources in their local area from the network operator. LTE-V2V is a variant of this specifically for automotive use cases ~\cite{MolinaMasegosa.2017}. This avoids the overhead of using the infrastructure as a relay but also has a cost in minimum association time and is dependent upon the network operator ceding resources on demand. Some proposals extend D2D cellular radio techniques into the unlicensed spectrum, or specifically, licensed sub-bands~\cite{Miao.2021}, however, this also has not been widely used in real world systems due to the recency of the specification and a dearth of capable hardware.


5G introduced the Ultra-Reliable Low Latency Communications (URLLC) service to address the issues with low latency medium access ~\cite{Feng.2021} in a centralized or decentralized manner, with a variety of proposed physical layer ~\cite{Le.2021} and mac layer ~\cite{Ali.2021} techniques. Despite the promise of these approaches, as well as 3GPP release 15 (including URLLC) being released in late 2017, roll out of these technologies into real world networks has been limited, and considerable research is ongoing as to the best implementation methods for 5G's technical goals~\cite{NavarroOrtiz.2020}.

\subsubsection{Hybrid Schemes}
Due to advancements in radio hardware, the most recent 802.11 revision, 802.11ax, has specified a physical radio resource allocation scheme that is far more similar to Cellular standards, with multiple fine-grained frequency divisions being available within a single logical channel. Many proposals have been made to have future Cellular standards directly inter-operate with Cellular networks to leverage the best capabilities of both~~\cite{Lagen.2020}. For robot networks, this may prove to be highly valuable, permitting human control overrides over Cellular infrastructure and low latency robot-to-robot communications with a unified logical network addressing system for easier lab-to-world deployment, and efficient operation in control schemes that have evolving requirements throughout a single deployment. Though exciting, these proposals are still in their nascent phase and many issues remain to be addressed.

Though the state of the art use of OFDMA in both 5G and 802.11ax significantly alleviate the contention problem due to the larger number of transmission slots made available, extremely dense ad-hoc robot networks may still run into the limit of CSMA/CA. Non-orthogonal media access (NOMA) is a very promising technology that has the possibility to further extend the effective simultaneous radio resources available ~\cite{DiZhang.2019}, and therefore reducing the contention problem. In cellular systems, the network operator still has authoritative control over their radio resources, and so, grant-based schemes induce significant overhead despite the reduced contention; though grant-free access has attracted significant attention ~\cite{Shahab.2020}. Even in grant-free schemes, NOMA still requires coordination to ensure that node configurations do not overlap, and hence, message loss due to contention remains an open problem in decentralized networks without a coordinating infrastructure.



\section{Communication-Aware Algorithms}
\label{sec:comms_aware}





Regardless of which underlying communication scheme or protocol is employed, unlimited and unconstrained communication cannot be assumed for any interactive scenario.
A significant amount of literature in multi-robot applications, however, has generally focused on designing control schemes that do not explicitly model this dependency.
This is reflected in the vast majority of literature in robot flocking \cite{soria2021distributed,zhou2021decentralized,tordesillas2021mader,luis2020online}.
We argue that the problem becomes more pronounced in cases where the robots need to deconflict and replan their motions in tight and constrained spaces \cite{zhou2021decentralized,tordesillas2021mader}.
While some consideration for communication asynchronicity is made in some of the more recent works \cite{tordesillas2021mader}, the challenge is generally far from being solved.

One straight-forward approach to handle this is to simply \textit{reduce} the amount of data (frequency, packet-size etc.) that needs to be communicated between agents.
In an exploration problem, this is often done through various novelty metrics that determine whether a new datapoint needs to be communicated \cite{kepler_approach_2020}.
Trawny \etal{} \cite{trawny_cooperative_2009} have proposed localization estimators that perform well by quantizing the transmitted information to very small packets, thereby tackling severe link constraints.

On the other hand, there is also a sustained research interest in modeling the communication channels between the agents, and factoring that as a constraint into the motion planning problem.
This is done primarily to ensure robustness of a control scheme against imperfect and noisy communications.
Alternatively, planning schemes have also considered communication as a sub-task (almost as if ``scheduling'' communications at intervals).
Finally, there are several approaches that consider a joint optimization scheme, where path planning and communication planning are carried out in tandem.
We divide this body of work into these three broad styles.


\subsection{Communication-Aware Planning}
\label{subsec:comm-aware-planning}
As mentioned earlier, planning robot motions or trajectories that consider some model of the underlying communication links is an active area of research.
Mularidharan and Mostofi provide a comprehensive overview of such methods \cite{muralidharan_communication-aware_2021}.
For instance, several authors have considered the task of coverage \& formation control by a team of robots.
Evidently, these domains require explicit factoring of communication constraints into the planning problem \cite{kepler_approach_2020}.
One way this is approached is by analyzing the stability of a formation under various communication link latencies \cite{zeng_joint_2019}.
This can then be then integrated into the control problem for a more reliable system \cite{zeng_wireless_2018, zeng_joint_2019} that is ``aware'' of such latencies.
Formation control laws are also explored that allow agents to maintain some degree of coordination while respecting limited communication ranges of their neighbors \cite{dimarogonas_bounded_2010,ji_distributed_2007}.
This approach is also sometimes utilized in the context of cooperative target localization \cite{stachura_cooperative_2011}, or under constraints of 3G/4G mobile networks \cite{olivieri_de_souza_coordinating_2015}.

Path planning has also been developed such that connectivity with a subset of base-stations \cite{mardani_communication-aware_2019}, or with some agents \cite{hsieh2008maintaining,schouwenaars2006multivehicle} is maintained.
Similar methods that plan for multiple robots (with connectivity constraints) involve using ACO-based (ant colony) planning \cite{fridman2008distributed} or genetic algorithms \cite{hayat2017multi}.

\subsection{Plan-Aware Communications}
In heavily constrained spaces, it is often desirable to design a network architecture that considers the planned path, and seeks opportunities to communicate therein.
Underwater robots, for instance, have very limited communication capacities \cite{doniec2010using,vasilescu2005data}, and this is an active field of interest; Zolich \etal{} \cite{zolich2019survey} provide a comprehensive survey on the various challenges and solutions.
Hollinger \textit{et. al.} have considered scheduling algorithms for underwater robotic sensor networks and show how path planning algorithms depend on these \cite{hollinger_communication_2011,anderson_communication_2021}.
Since bandwidth and interference constraints are much more severe in these environments, such scheduling algorithms often model the \textit{value} of communicating at a particular timestep \cite{anderson_communication_2021,best_planning-aware_2018}.
This also plays a role in determining whether communicating has a positive impact on the state of the robot system \cite{alshehri_modeling_2021,unhelkar2016contact}, and is also studied as an online decision problem \cite{tsiogkas_towards_2019}, and an optimization problem that considers when/what to communicate \cite{marcotte2020optimizing}.

\subsection{Joint Planning}
Several of the works listed in the previous subsections may also be seen as jointly optimizing for communication quality as well as path qualities.
However, there are other approaches that attempt to explicitly model this optimization problem.
For instance, Kantaros and Zavlanos \cite{kantaros_distributed_2016} propose a scheme that \textit{alternates} between the two optimization problems sequentially.
The nature of this scheme often makes it difficult to prove hard guarantees regarding optimality; however, a more hybrid approach in which the two controllers interact can offer more guarantees on network integrity available data-rates \cite{zavlanos_network_2013}.
A joint optimization scheme, on the other hand, can formulate this problem well; for instance, using an LQ (linear-quadratic) form can additionally offer robustness guarantees as well \cite{kassir_decentralised_2012}.
Yet another means of joint optimization is to consider the system as a cyber-physical system (CPS), where the `cyber' controller handles the communications domain, and the `physical' controller handles the kinematics of the robot \cite{fink_robust_2012}.
Such models allow designers to factor various other elements of a CPS system, such as dynamically adapting one of the subsystems (communication capacity) while still maintaining the coupling with the other \cite{le_ny_adaptive_2012,stephan_concurrent_2017}.


\section{Leveraging Machine Learning for Communication}
\label{sec:learning}


Designing bespoke, handcrafted communication protocols and behaviors is tedious and difficult. Firstly, numerous works point to the hardness of synthesizing decentralized policies (that have to operate in a partially observable regime), even when a centralized template is known~\cite{halsted_Survey_2021, amato2013decentralized}, and they leave the question of \textit{how} (what, when, and to whom) to communicate unanswered. Secondly, the vast majority of existing robot communication strategies are based on idealistic operational assumptions, and besides a few specialized approaches to dealing with message loss, delay, or corruption, e.g., ~\cite{gil_Guaranteeing_2017, saulnier_Resilient_2017a, parker_ALLIANCE_1998}, it is not at all clear how to approach such problems in a manner that is transferable across applications. Leveraging machine learning methods is a promising new avenue to tackle some of these challenges.

\subsection{Learning Communication Mechanisms}
Message routing decisions in robotic mesh networks are complicated by highly dynamic topologies. While many routing mechanisms exist in ad-hoc networks, these generally depend upon relatively slowly changing network conditions to function effectively. Many manually specified heuristic methods exist~\cite{ShumeyeLakew.2020}, however these may lead to sub optimal decisions as they may be constructed upon incorrect assumptions about the target network environment. Learning methods provide an attractive alternative, and have been explored in some depth in routing generally ~\cite{Mammeri.2019}. An interesting example in the context of FANETs can be found in Zheng et al. ~\cite{Zheng.2018}, who propose RLSRP which applies an online reinforcement learning method to the routing decision problem and shows improved performance across several metrics, including delivery latency.

Channel modeling and resource allocation are also key networking problems that are challenging for first principles methods to solve that can be improved with learning ~\cite{Bithas.2019}. Unsupervised learning has be applied to channel modeling, which allows for the optimization of transmission power by accurately estimating the quality of links to other network participants ~\cite{Wang.2019c}.

\subsection{Learning Communication Behaviors}

Learning-based methods have proven effective at designing robot control policies for an increasing number of tasks~\cite{rajeswaran_Generalization_2017, tobin_Domain_2017}. Recent work utilizes a data-driven approach to solve multi-robot problems, for example for multi-robot motion planning in the continuous domain~\cite{everett_Motion_2018} or path finding in the discrete domain~\cite{sartoretti_PRIMAL_2019}.

Yet, research on learning how to \textit{synthesize robot-to-robot communication policies} is nascent. From the point of view of an individual robot, its local decision-making system is incomplete, since other agents' unobservable states affect future values. While the manner in which information is shared is crucial to the system's performance, the problem is not well addressed by hand-crafted (bespoke) approaches. Learning-based methods, instead, promise to find solutions that balance optimality and real-world efficiency, by bridging the gap between the qualities of full-information centralized approaches and partial-information decentralized approaches~\cite{prorok2021holy}. 

Key to the decentralization of centralized (optimal) policies is the property of \textit{permutation equivariance}. Permutation equivariance ensures that at the robot network level, the set of actions automatically rearranges itself as the agents swap order. One of the earliest works that satisfies this property is~\cite{paulos2019decentralization}. This was concurrently developed by a line of work that builds on Graph Neural Networks (GNNs), which are permutation equivariant by design~\cite{scarselli_graph_2009, Gama19-Architectures, prorok_Graph_2018}. GNNs have since then shown promising results in learning explicit communication strategies that enable complex multi-agent coordination~\cite{modgnn,khan_2020, tolstaya_Learning_2019a, li_Graph_2020a}. 

When deploying GNNs in the context of multi-robot systems, individual robots are modeled as nodes, the communication links between them as edges, and the internal state of each robot as graph signals. By sending messages over the communication links, each robot in the graph indirectly receives access to the global state. A key attribute of GNNs is that they compress data as it flows through the communication graph. 
In effect, this compresses the global state, affording agents access to relevant encodings of global data. 
Since encodings are performed locally (with parameters that can be shared across the entire graph), the policies are intrinsically decentralized.
In cases where the downstream task is tightly coupled with the communication requirements, it is beneficial to optimize the communication strategy jointly with perception and action policies. This was done in~\cite{hu_scalable_2022}, for multi-robot flocking, and in~\cite{li_Graph_2020a}, for multi-agent path planning. These frameworks implement a cascade of a convolutional neural network (CNN) and a GNN, which they jointly train so that image features and communication messages are learned in conjunction to better address the specific task.
Recent work also shows how GNNs can be augmented by \textit{attention modules} to produce \textit{message-aware} communication strategies that allow robots to discern between important and less important message elements~\cite{li_Messageaware_2021}. 

Approaches from within the multi-agent reinforcement learning (MARL) community tackle the learning of continuous communication protocols by formulating the problem as a Decentralized Partially Observable Markov Decision Process (Dec-POMDP)~\cite{sukhbaatar_learning_2016, singh2018learning, jiang_learning_2018}. The work in~\cite{das_tarmac:_2019} learns a targeted multi-agent communication strategy by exploiting a signature-based soft attention mechanism (whereby \textit{message relevance} is learned). Similarly, the work in~\cite{serra-gomez_whom_2020} has each robot learn to reason about other robots’ states and to more efficiently communicate trajectory information (i.e., when and to whom), and applies the solution to the problem of collision avoidance. While efficient \textit{cooperative} communication strategies are desirable, the work in~\cite{blumenkamp_Emergence_2020b} shows how separate robot teams can learn to communicate with \textit{adversarial} strategies that contribute to manipulative (non-cooperative) behaviors. Clearly, underlying training paradigms need to be carefully designed to avoid such outcomes.





\section{Challenges and Open Problems}
\label{sec:future}



We finally present some avenues of research and engineering that are worth exploring in order to address our critiques discussed so far.
We categorize them into four broad \textit{Open Problems}.

\textbf{1. Co-design.}
An emergent theme throughout this survey is the lack of approaches that co-design the robot and its communication capabilities.
A variant of this concept \cite{mechraoui2009co,mechraoui2011co} considers a basic parallel reconfiguration of a network as well as the robot's controller that can be beneficial when the robot moves across network stations. 
However, a true co-design scheme will jointly evolve all layers of the networking stack to favor the robotic task at-hand.
Design of a meta-system that is able to compute the limitations of robotic requirements as well as network capabilities and dynamically throttle both may be essential to safe deployment of robots into the real world.
Any robotic control algorithm that uses explicit communications is vulnerable to failure if the network unexpectedly under-delivers, and performs sub-optimally if the network over-delivers -- managing this resource allocation problem in a real world multi-robot setting is a subject we will tackle in our future work.

\textbf{2. Data-driven optimization.}
Machine learning, and specifically, reinforcement learning, can drive the development of multi-robot communications into new and interesting paradigms.
Existing approaches that already learn what/when to send (and whom to send to) \cite{serra-gomez_whom_2020, li_Messageaware_2021, paulos2019decentralization} still often depend on hand-designed architectures and specific task groups.
With sufficiently large datasets, novel machine-learning architectures also have the potential to learn to optimize multiple aspects of multi-robot systems at once (e.g., perception, action and communication \cite{hu_scalable_2022}).

\textbf{3. Sim-to-real of robot networks.}
The problems in sim-to-real transfer of robot coordination strategies are generally exacerbated by the ``reality gap" found in communications~\cite{prorok2021holy}.
Practical communication links suffer from message dropouts, asynchronous and out-of-order reception, and decentralized mesh topologies that may not offer reliability guarantees.
Since multi-robot policies are typically trained in a synchronous fashion, these factors are hard to capture and simulate~\cite{blumenkamp2021framework}.
Furthermore, very few studies have captured any of these network effects in a large-scale setting \cite{gielis2021improving}.
Consequently, we find that embedding the reality gap of robot networking into data-driven approaches to multi-robot planning is an open research domain.

\textbf{4. New technologies/schemes.}
As discussed in Section~\ref{subsec:comms_protocols}, there is a need for wireless data standards that specifically target the communication requirements of connected robots.
The \textsc{IEEE 1920} working group is a significant step in this direction, which was formed to propose a protocol that is intended for autonomous robotic networks~\cite{IEEE.2022}.
Such a protocol is likely to be founded on 802.11bd since it is already a significant leap forward \cite{Naik.2019} from the legacy 802.11p standard used in V2V standards today.

Additionally, future 5G updates and 6G cellular communications promise dramatic improvements that hold the potential to bring cloud- and edge-computing at the forefront of many data-intensive multi-robot collaborations.
Finally, we also note that geographic routing in FANETs may be an enabling technology for practically dealing with highly dynamic routing topologies.
This will, however, require holistic developments in robot control algorithms that work in tandem to avoid an additional information distribution problem.


\section{Conclusion}
\label{sec:conclusion}
Through this manuscript, we have presented a survey of communication technologies and their role in enabling multi-robot applications.
We have broadly covered the various technologies that have played key roles in networked robotics, and have also discussed how state-of-the-art robot applications typically deal with network constraints.
Our approach to this has been mostly critical, and thus, has identified several deficiencies in the way robotics and networks have evolved.
Towards the end, we also cover machine learning approaches and their role in developing data-driven communication strategies.
We conclude the article with a list of challenges and open problems that the community currently faces, and also provide an outlook for how learning-based approaches can tackle several of them.

\section{Acknowledgments}
\label{sec:acks}
Jennifer Gielis was supported by an EPSRC Doctoral Training studentship. The authors also gratefully acknowledge support from  ARL DCIST CRA W911NF-17-2-0181 and the European Research Council (ERC) Project 949940 (gAIa).

\subsection*{Declarations}
\textbf{Conflict of interest.}
The authors declare that they have no conflicts of interest.
\textbf{Human and Animal Rights, and Informed Consent.}
This article does not contain any studies with human or animal subjects performed by any of the authors.




\printbibliography[title=References]

\end{document}